\documentclass[letterpaper,10pt,conference]{ieeeconf}  % Comment this line out if you need a4paper

\IEEEoverridecommandlockouts                              % This command is only needed if 
\usepackage{graphicx} % for pdf, bitmapped graphics files
\usepackage{epsfig} % for postscript graphics files
\usepackage{mathptmx} % assumes new font selection scheme installed
\usepackage{times} % assumes new font selection scheme installed
\usepackage{amsmath} % assumes amsmath package installed
\usepackage{amssymb}  % assumes amsmath package installed
\usepackage{subfigure}
\usepackage{algorithm}
\usepackage{algpseudocode}

\newtheorem{problem}{Problem}

\graphicspath{{C:/Users/Aamodh/Documents/BCI/Documentation/plots/EOG/U_D_Detection/},{C:/Users/Aamodh/Documents/BCI/Documentation/plots/EOG/L_R_Detection/},{C:/Users/Aamodh/Documents/BCI/Documentation/plots/HMM/}}

\title{\LARGE Brain-Swarm Interface (BSI): Controlling a Swarm of
  Robots\\ with Brain and Eye Signals from an EEG Headset}
\author{Aamodh Suresh$^{1}$ and Mac Schwager$^{2}$% <-this % stops a space
\thanks{$^{1}$A.~Suresh is with the Department of Mechanical Engineering, Boston University, Boston, MA 02215, USA {\tt\small aamodh@bu.edu}}%
\thanks{$^{2}$M.~Schwager is with the Department of Aeronautics and Astronautics, Stanford University, Stanford, CA 94305, USA {\tt\small schwager@stanford.edu}}%
\thanks{This work was supported in part by NSF grants IIS-1350904 and  CNS-1330036.  We are grateful for this support.}
}

\DeclareMathOperator*{\argmax}{\arg\!\max}
\begin{document}

\maketitle
\thispagestyle{empty}
\pagestyle{empty}

%%%%%%%%%%%%%%%%%%%%%%%%%%%%%%%%%%%%%%%%%%%%%%%%%%%%%%%%%%%%%%%%%%%%%%%%%%%%%%%%
\begin{abstract}
  This work presents a novel marriage of Swarm Robotics and Brain
  Computer Interface technology to produce an interface which connects
  a user to a swarm of robots.  The proposed interface enables the
  user to control the swarm's size and motion employing just thoughts
  and eye movements. The thoughts and eye movements are recorded as
  electrical signals from the scalp by an off-the-shelf
  Electroencephalogram (EEG) headset. Signal processing techniques are
  used to filter out noise and decode the user's eye movements from
  raw signals, while a Hidden Markov Model technique is employed to
  decipher the user's thoughts from filtered signals. The dynamics of
  the robots are controlled using a swarm controller based on
  potential fields.  The shape and motion parameters of the potential
  fields are modulated by the human user through the brain-swarm
  interface to move the robots.  The method is demonstrated
  experimentally with a human controlling a swarm of three M3pi robots
  in a laboratory environment, as well as controlling a swarm of 128
  robots in a computer simulation.
\end{abstract}

%%%%%%%%%%%%%%%%%%%%%%%%%%%%%%%%%%%%%%%%%%%%%%%%%%%%%%%%%%%%%%%%%%%%%%%%%%%%%%%%

\section{INTRODUCTION}

%Basic concepts of BCI and hybrid BCI\\
In this paper we present a new brain machine interface for a human
user to control a swarm of robots, which we call a Brain-Swarm
Interface (BSI).  The BSI uses an off-the-shelf Electroencephalogram
(EEG) headset to record brain and muscle activity from the user's
scalp.  We use both signals from brain neuron activity, as well as
signals from contracting muscles due to eye movement, to generate
control signals for the robot swarm.  We allow the user to control the
dispersion/aggregation of the swarm, as well as the direction of
motion of the swarm.  The dispersion/aggregation is determined from
the user's brain neuronal signals, and is decoded using an HMM-based
method.  The direction of motion is decoded from the user's eye
movements with a multi-step signal processing algorithm.  The robots
maintain a cohesive swarm using a potential-field based swarm
controller, and the dispersion/aggregation and direction of motion
commands influence the motion of the robots through parameters in
their swarm controller.

Brain Computer Interfaces hold great promise for enabling people with
various forms of disabilities, from restricted motion due to injury or
old age, to severe disabilities like the ALS (Locked-in syndrome),
Tetraplegia, and paralysis.  Several works have investigated using
BCIs for controlling prosthetics \cite{wheelchair},
\cite{robotic_arm}, and for medical rehabilitation.  If implemented
effectively, BCI technology may allow people with disabilities the
power to manipulate their environment, and to move themselves within
their environment \cite{ALS_tetraplegia}.  Researchers have also
developed brain interfaces for controlling single mobile robot
platforms. For example Bin He et al. \cite{BCI_quadcopter} have
demonstrated 3D control of a quadcopter, and Tim Bredtl et
al. \cite{quad_tele} have remotely teleoperated a UAV, both using
motor imagery BCI.  C J Bell et al. \cite{humanoid} have controlled a
humanoid with a non invasive BCI using P300 signals.  However, brain
computer interfaces for controlling swarms have received little
attention in the literature.

Whereas the motivation for a BCI operated prosthetic or wheelchair is
evident, the applications for a brain swarm interface may be less
obvious.  We envision several applications for this technology.
Firstly, people who are mobility-impaired may use a swarm of robots to
manipulate their environment using a brain-swarm interface.  Indeed, a
swarm of robots may offer a greater range of possibilites for
manipulation than what is afforded by a single mobile robot or
manipulator.  For example a swarm can reconfigure to suit different
sizes or shapes of objects, or to split up and deal with multiple
manipulation tasks at once. Another motivation for our work is that
using a brain interface may unlock a new, a more flexible, way for
people to interact with swarms.  Currently human swarm interfaces are
largely restricted to gaming joysticks with limited degrees of
freedom.  However, swarms typically have many degrees of freedom, and
the brain has an enormous potential to influence those degrees of
freedom beyond the confines of a traditional joystick.  We envision
that the brain can eventually craft shapes and sizes for swarm, split
a swarm into sub swarms, aggregate or disperse the swarm, and perhaps
much more.  In this work, we take a small step toward this vision.

% Traditionally, BCIs used sensor arrays implanted in the brain or
% limbs.  Because of the surgical procedures required, this work was
% largely limited to the field of neuro-prosthetics and rehabilitation.
% With the evolution of no- invasive sensing methods such as EEGs, which
% do not require any medical procedures, BCI research is growing in non
% clinical fields such as robotics.

% A Brain Computer Interfac (BCI), also known as a Brain Machine
% Interface (BMI), is a technology which aims to build bridges between
% the brain and an external device to serve as a communications
% pathway. BCIs use a device to measure and analyze brain activity, and
% convert it into interpretable signals to control an external device.
% A common non-invasive device used in BCIs to measure brain activity is
% an which detects electrical signals from the scalp arising from a
% myriad of neurons firing synchronously.

\begin{figure}
\centering
\includegraphics[width=\linewidth]{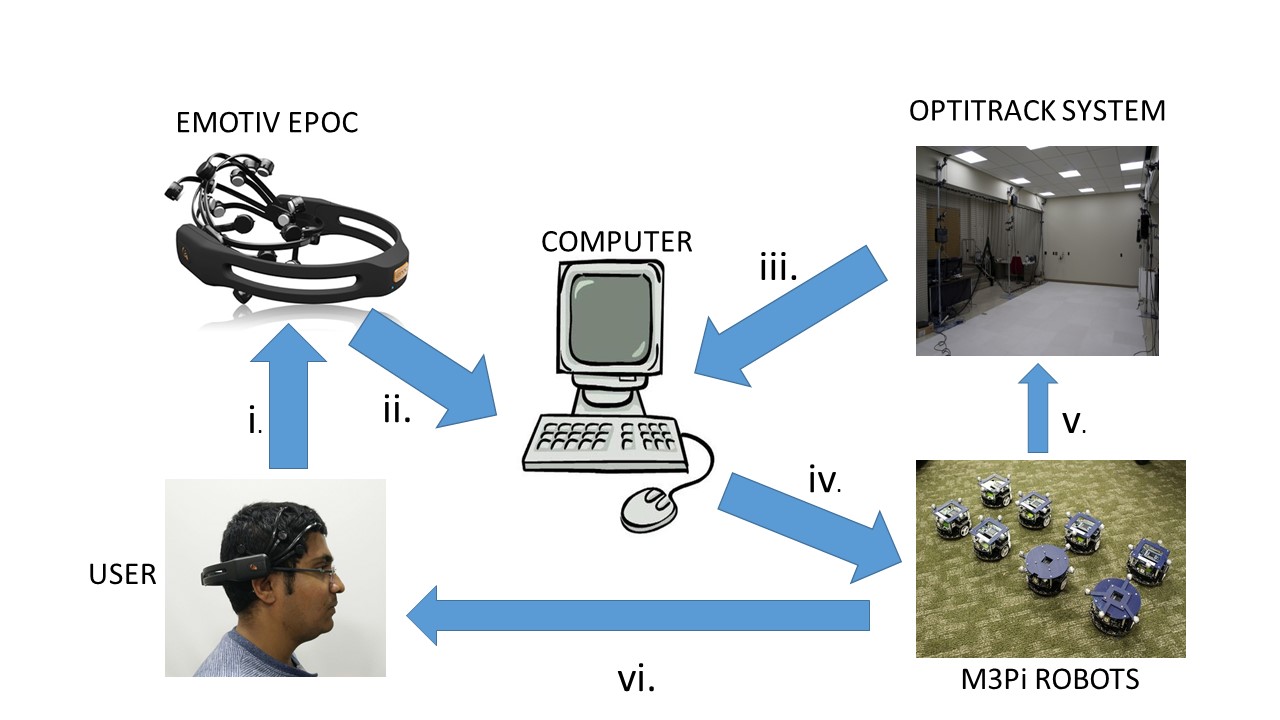}
\caption{Key components the system and their interaction. $i).$ The
  user generates brain and eye movement signals. $ii).$ The Emotiv
  Epoc headset records these signals from the scalp and wirelessly
  transmits them to the computer. $iii).$ The computer receives the
  pose of the robots. $iv).$ The computer decodes the eye movements
  and thoughts, and transmits control signals to the robots. $v).$ The
  Optitrack captures pose of robots. $vi).$ The user receives visual
  feedback.}
\label{control_loop}
\end{figure}

The brain is arguably the most complex organ in the human body. Brain
activity occurs mainly in the chemical and electromagnetic
domain. Different areas of the brain are responsible for different
thoughts and actions. The EEG is one of the few ways in which brain
activity can be monitored in a non-invasive manner. An EEG measures
the electrical signals generated by the brain, but it also measures
electrical noise and artifacts generated by eye movements and other
muscle activity.  % BCI by definition involves in processing only pure
% brain signals after removing artifacts and noise.
We use the concept of Hybrid BCI or hBCI popularized by Pfurtscheller
et al. \cite{hBCI} and utilize the "artifacts" generated by eye
movements, which are usually discarded in BCI systems, as control
input to modulate the position and size of a robotic swarm.  A survey
of other hBCI systems, which do not include application to robotic
swarms, can be found in \cite{hBCI_survey}.  

One of the key components of a typical BCI paradigm is the machine
learning and pattern recognition pipeline. A survey by F Lotte et
al. \cite{classifier_survey} surveys the various types of classifiers
that have been used in the literature.  In our work, we adopt a Hidden
Markov Model (HMM) technique to estimate the user's thought state, as
the dynamic nature of the technique is well-suited to online
estimation and control input generation.  We use outputs from an
Emotiv Epoc EEG headset as observations for the HMM algorithm.  The
HMM algorithm then maps these continuous observations to discrete
thought states which in turn generate the control inputs for the
swarm. The control inputs are converted to motor speeds for the
respective robots and transmitted through zigbee protocol
wirelessly. The user receives visual feedback from the ground robots,
either by directly observing the robots, or from a live visual feed
rendered on a video screen.  The main interacting components in our
system are illustrated in Fig. \ref{control_loop}.  

The rest of this paper is organized as follows.  In Section
\ref{Sec:ProblemFormulation} we introduce notation and formally state
the problem.  In Section \ref{Sec:System} we describe the signal
processing and control components that go into our system.  Section
\ref{Sec:Experiments} presents the results of simulation and hardware
experiments, and conclusions are given in
Section~\ref{Sec:Conclusions}.

\section{PROBLEM FORMULATION}
\label{Sec:ProblemFormulation}
% Brain signals are very high dimensional due to billions of individual
% neurons firing. However, because of technological limitations,
% non-invasive EEG headsets in general have sensors which are only
% capable of measuring large scale neurons firing. 
Let the position of the swarm be described by the vector $x(t) =
[x_1^T(t), \ldots, x_M^T(t)]^T \in \mathbb{R}^{nM}$ where $M$ denotes
the number of agents in an $n$ dimensional Euclidean space.  The
position of the $i^{th}$ individual at time $t$ is described by
$x_i(t) \in \mathbb {R}^n$.  Suppose the user has an intended
trajectory for the swarm, which we denote $x^*(t)$.  When thinking
about this trajectory, the user's EEG headset produces signals at time
$t$, denoted by $U(t)\in\mathbb{R}^{N}$, where $N$ is the number of
sensors on the EEG headset.  Let $\Theta$ be a vector of control
parameters of the system given by $\Theta(t)=(a(t),b(t),v(t))$ where
$a(t)$, $b(t)$ $\in\mathbb{R}$ denote the attraction and repulsion
gains, respectively, which are used to control the size of the swarm,
and $v(t)\in \mathbb{R}^n$ denotes the displacement vector for
controlling the movement of the swarm.  We state our problem as
follows.
\begin{problem}[Brain-Swarm Interface]
\label{Prob:BSI}
Design a signal processing pipeline to determine the swarm control
parameters $\Theta(t)$ from the EEG signals $U(t)$, so that the swarm
trajectory approaches the user's intended swarm trajectory,
$x(t)\rightarrow x^*(t)$.
\end{problem}

To control the swarm, we adapt a potential field based swarm
controller of a type that is common in controlling swarms of ground
robots
\cite{Gazi&Passino,Olfati-Saber,HowardEtAlDARS02PotentialFieldSensorNetDeployment}.
With this type of controller, the system dynamics are given by
\begin{equation}
\dot{x_i}= \sum_{j=1,j\neq0}^Mf_{a,b}(x_i,x_j)+v,
\end{equation} 
where $i=1,2, \ldots, M$ and $f_{ij}:\mathbb {R}^n \to \mathbb {R}^n$
is a function which depends on pairwise interactions between agents
$i$ and $j$, and has parameters $a$, $b$, and
$v$. % Let $g(\cdot)$ be defined by
% \begin{equation}
%   g(y)=-(ag_a(\|y\|)+bg_r(\|y)\|))\frac{y}{\|y\|},
% \label{force_function_eqn}
% \end{equation}  
% where $g_a:\mathbb {R}^+ \to \mathbb {R}^+$ represents the attraction
% component, $g_r:\mathbb {R}^+ \to \mathbb {R}^+$ represents the
% repulsion component, and $\|y\|$ is the euclidean norm.

In our proposed solution, we first train an HMM, and use it to
determine the ``thought state'' of the user $Q(t)\in\{1,2\}$, where
for example, $Q = 1$ indicates ``aggregation'' and $Q = 2$ indicates
``dispersion.''  Then we map this though state $Q_t$ to values for
$\{a(t),b(t)\}$ to control the aggregation/dispersion of the swarm.
Secondly, we design an eye movement classifier which takes EEG signals
$u(t) \subset U(t)$, where $u(t)\in\mathbb{R}^k$ with $k<N$, as input
to determine the user's intended motion of the swarm, producing a
control parameter $v(t)$ for the swarm.  This pipeline is shown
graphically in Fig.~\ref{Control_flow}, and its components are
described in more detail in the following section.

\section{SYSTEM DESCRIPTION}
\label{Sec:System}
\begin{figure}
\centering
\includegraphics[width=3.5in,height=4in,keepaspectratio]{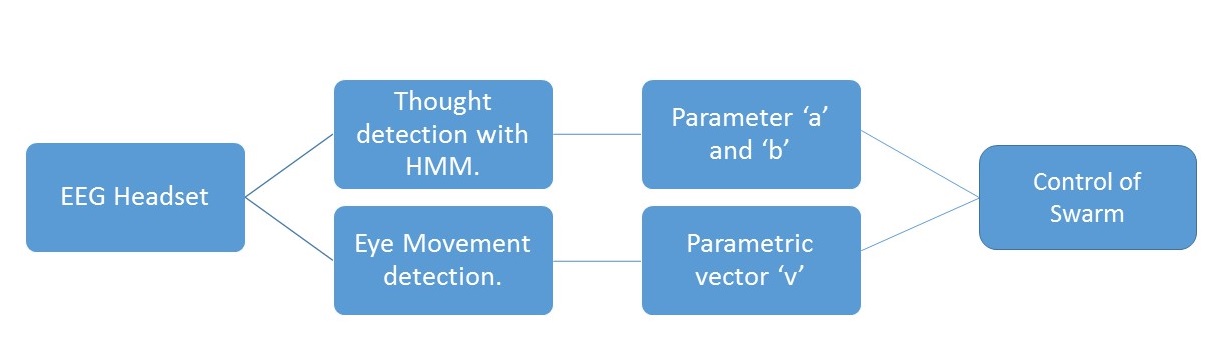}
\caption{Process Flow from left to right.}
\label{Control_flow}
\end{figure}

In the previous section we stated the problem, and introduced the main
elements of our solution as shown in Fig \ref{Control_flow}. The
following paragraphs will describe the HMM, eye movement detection,
and formation control strategies in detail.

\subsection{Decoding Thoughts Using an HMM}
\label{Sec:HMM}
Based on a survey by F. Lotte et al. \cite{classifier_survey} we
adopted Hidden Markov Models to train and classify EEG
data. Previously, researchers have used HMMs for BCIs. Pfurtscheller
et al. \cite{HMM_Pfurtscheller} used an HMM to classify EEG data, and
HMMs have been used in conjunction with other techniques
\cite{pca_svm_hmm}, as well. Our novel HMM implementation uses
performance metrics generated by the Emotiv software suite as
observations and maps them to discrete thoughts. We observe that each
thought will corresponds to different signatures of these metrics.
However, these signatures vary greatly across different trials and
experimental conditions, so a heuristic approach to classify them is
not effective.

Instead we implement a training phase to train the HMM to detect the
users though signatures and transition probabilities.  Using standard
terminology and notation for HMMs, \cite{Rabiner}, \cite{EMalgo}, here
we will describe our system in detail.

\subsubsection{HMM: An Overview}
A Hidden Markov Model is a joint probabilistic model of a collection
of discrete random variables ${O_1,...,O_T,Q_1,...,Q_T}$ described by:
	  \begin{equation}
	  P(O,Q)=P(Q_0)\prod^T_{t=1}P(Q_t|Q_{t-1})\prod^T_{t=0}P(O_t|Q_{t}) \label{HMM_equation}
\end{equation}	   
Many algorithms exist for both learning the transition probabilities
and observation probabilities of such a system from data, and
determining a likely sequence of states hidden $\{Q_t\}$ from data.
The observations of these states $O_t\in\mathbb{R}^l$ can be discrete
or continuous.

The main components of the model from Equation \ref{HMM_equation} are:
\begin{itemize}
\item $P(Q_0)$ which is the initial state probability distribution and
  is represented by $\pi_i=P(Q_1=i)$ where $i\in\{1,...,m\}$,
\item ${P(Q_t|Q_{t-1})}$ is the state transition probability
  represented by the matrix $X =\{x_{ij}\}=P(Q_t=j|Q_{t-1}=i)$ where
  $i,j\in\{1,...,m\}$
\item ${P(O_t|Q_{t})}$ is the observation probability represented by
  the gaussian distribution
  $B_i(o_t)=P(O_t=o_t|Q_t=i)=\frac{\exp\{-1/2(o_t-\mu_i)^TR_i^{-1}(o_t-\mu_i)\}}{\det\sqrt{2\pi
      R_i}}$ with means $\mu_i\in\mathbb{R}^l $ and covariance matrix
  $R=\{r_{i,j}\}\in\mathbb{R}^{l\times l}$.
\end{itemize}
The model can be completely described by parameters
$\theta=\{\pi_i,X,\mu_i,R\}$. The initial phase involves learning
these parameters $\theta$ to generate the model using training data
consisting of observations of the expected state space. We employ the
Baum-Welch algorithm (a version of Expectation Maximization (EM)) to
train the model parameters. After the training phase we use the
Forward Algorithm to estimate the state of the model online using the
current observations.

We are use a two state HMM which represents two distinct thoughts of
the user, so $m=2$ in our case. The observation space consists of the
EEG output $U(t)$.  In this case, the signal $U(t)$ is derived from
performance metrics provided by the manufacturer of the EEG headset,
Emotiv.  Emotiv provides six metrics: ``Engagement", ``Meditation",
``Excitement", ``Frustration", ``Valence" and ``Long-Term Excitement"
out of which we use the first three metrics, which makes $l=3$ and
also $O_t\in [0,1]$ for all metrics.

\subsubsection{The Training Phase with Baum-Welch Algorithm}
Training data consisting of the three metrics mentioned before is
recorded in a single trial. During the training period the user
repeats two thoughts through a pre-defined switching sequence.  We
enforced this thought sequence using a timed slide presentation which
the user observed during training.  This training data is fed into the
Baum-Welch algorithm which estimates the parameters $\theta$ for the
system.

The EM Algorithm is a two step iterative process (Expectation followed
by Maximization), which can be expressed in the single expression
 \begin{equation}
 \theta_{c+1}=\argmax_\theta E_{P(Q,O|\theta_c)}[\mathcal{L}(Q,\theta)],
 \label{EM_eqn}
 \end{equation}
 where $\mathcal{L}(Q,\theta)=\ln P(Q,O|\theta)$ is the log likelihood
 function, $Q$ is the unobserved state, $\theta$ is the unknown
 parameters of the model, and $O$ is the observed variable. The
 expectation step can be summarized by calculating the following
 quantities:
 \begin{equation}
   \alpha_i(t)=P(O_1=o_1,...,O_t=o_t,Q_t=i|\theta),
 \label{alpha_eqn}
 \end{equation}
 where $\alpha_i(t)$ is known as the forward variable, and is the
 probability of ending in state $i$ and seeing the partial
 observations $\{o_1,...,o_t\}$ given the model parameters $\theta$,
 and
\begin{equation}
  \beta_i(t)=P(O_{t+1}=o_{t+1},...,O_T=o_T|Q_t=i,\theta),
\end{equation} 
where $\beta_i(t)$ is known as the backward variable, and is the
probability of observing partial sequences $\{o_{t+1},...,o_T\}$ given
the model parameters and state at time $t$.  With $\alpha$ and $\beta$
we can compute
 \begin{equation}
   \gamma_i(t)=P(Q_t=i|O,\theta)=\frac{\alpha_i(t)\beta_i(t)}{\sum^m_{j=1}\alpha_j(t)\beta_j(t)},
 \end{equation}
 where $\gamma_i(t)$ is the probability of being in state $i$ given
 the model parameters and observations, as well as
\begin{equation}
\begin{split}
\zeta_{ij}(t) &=\frac{P(Q_t=i,Q_{t+1}=j,O|\theta)}{P(O|\theta)}  \\
&=\frac{\alpha_i(t)x_{ij}B_j(o_{t+1})\beta_j(t+1)}{\sum_{i=1}^m\sum_{j=1}^m\alpha_i(t)x_{ij}B_j(o_{t+1})\beta_j(t+1)},
\end{split}
\end{equation} 
where $\zeta_{ij}(t)$ is the probability of being in state $i$ at time
$t$ and state $j$ at time $t+1$.

Using the above defined quantities in the expectation step we can
estimate the parameters $\theta$ in the maximization step of our
system as follows :
\begin{center} 
\begin{gather}
\mu_i^{p+1}=\frac{\sum_{t=0}^T\gamma_i^p(t)o_t}{\sum_{t=0}^T\gamma_i^p(t)} \\
R_i^{p+1}=\frac{\sum_{t=0}^T\gamma_i^p(t)(o_t-\mu_i^{p+1})(o_t-\mu_i^{p+1})^T}{\sum_{t=0}^T\gamma_i^p(t)} \\
X_T^{p+1}=\frac{\sum^T_{t+1}\zeta_{ij}^p(t-1)}{\sum_{t=1}^T\gamma_j^p(t-1)}\\
\pi_i^{p+1}=\gamma_i^p(0).
\label{Mstep_eqns}
\end{gather}
\end{center}

Naturally, for this procedure to start we need $\theta_0$ which is the
set of initial model parameters from which recursion begins according
to Eqn. (\ref{EM_eqn}).  We adopt a K-means clustering approach to
initialize the model parameters, specifically the mean matrix
$\mu$. We used a two class K-means approach with the observations $O$
as inputs, which give us the centroids of the two classes that we used
to initialize $\mu$. The other parameters of $\theta$ are initialized
randomly. We stop the iterative process when we observe only minute
changes (order of $10^{-4}$) in the estimated parameters from their
previous estimations.
\begin{figure}
\centering
\subfigure[Observation data over training period\label{observation_data}]{\includegraphics[width=0.50\textwidth]{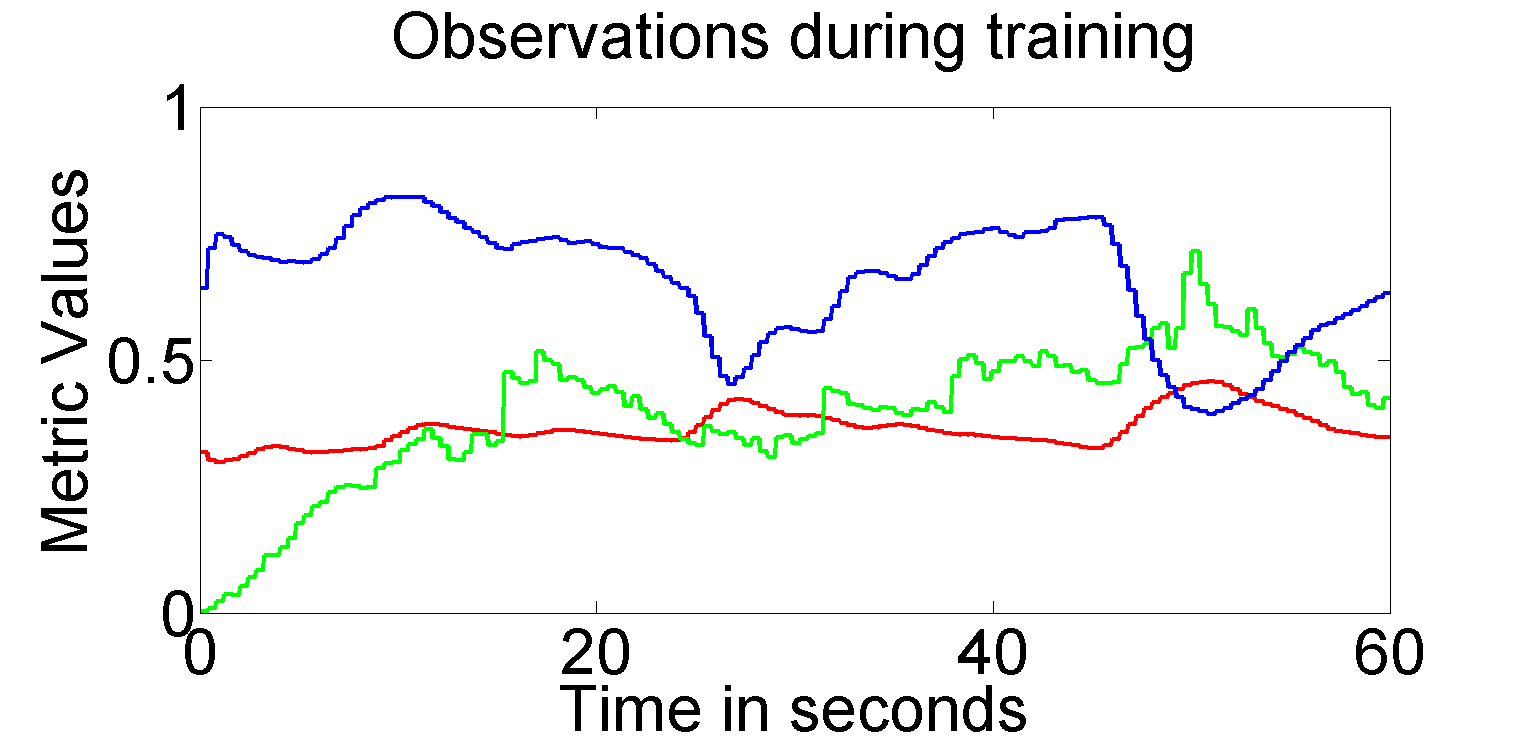}}
\subfigure[State estimation of the training data\label{state_estimation}]{\includegraphics[width=0.50\textwidth]{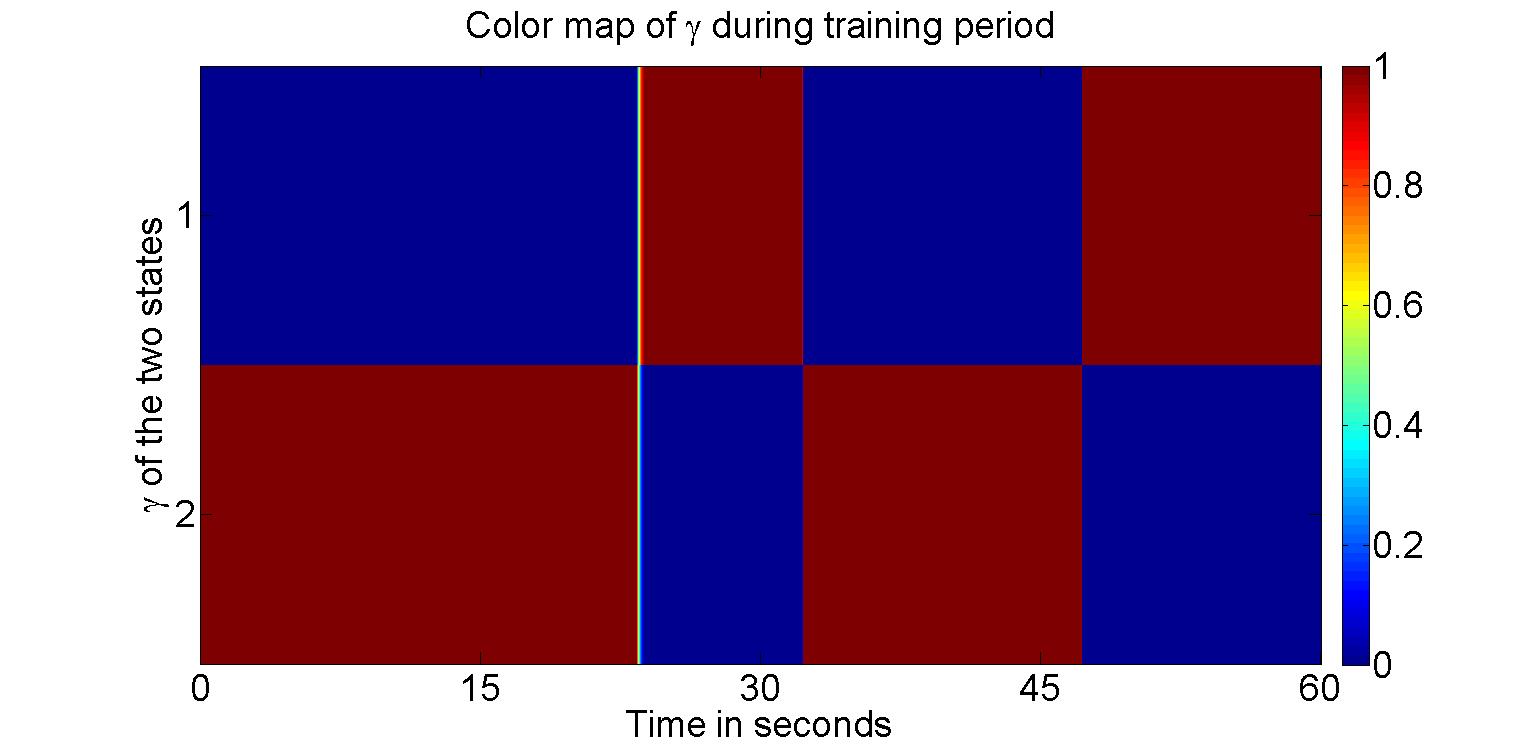}}
\caption{(a) Shows the 3 metrics from the EEG signal (red for
  'Meditation', green for 'Excitement' and blue for 'Engagement')
  during the training period of 60 seconds.  (b) Shows the estimated
  state of the HMM during the training period by plotting the color
  map of $\gamma$.  The lower state corresponds to a ``disperse''
  thought by the user, and the upper to an ``aggregate'' thought by
  the user. \label{Fig:Baum_welch}}
 \end{figure}
 A typical training signal from our experiments is shown in
 Fig. \ref{observation_data} and the resulting state sequence after
 training is shown in Fig. \ref{state_estimation}. The user visited
 the two states (thoughts) twice each during the training period. From
 Fig. \ref{state_estimation} one can see that the Baum-Welch algorithm
 detects the switching sequence between the thoughts, since at each
 all times the state is found to be decisively in either one or the
 other state with high probability.  That is, at all times one state
 is red (meaning the probability that the user is in that thought
 state is nearly one), while the other is blue (meaning probability
 that the user is in that thought state is almost zero).

\subsubsection{Online Estimation with the Forward Algorithm}
After the model parameters $\theta$ have been estimated we can employ
the HMM for online estimation of the state.  However, we cannot
immediately proceed to online estimation.  First, we have to relate
the HMM states back to the original thoughts.  Specifically, we do not
know whether $Q = 1$ means an ``aggregate'' thought and $Q = 2$ means
a ``disperse'' thought, or visa versa.  We assign the abstract states
to meaningful thoughts by examining the order of thoughts visited by
the user during training to the $\gamma(t)$ value calculated during
the training.  Which ever assignment makes the $\gamma$ sequence best
match the thought sequence is the chosen assignment.

Online estimation of state is now a straightforward application of the
Forward algorithm for HMMs using the learn parameters from the
training phase.  This allows us to find the most likely sequence of
thought states on line as a streaming signal arrives from the EEG.
All we need for this phase is to calculate the value of $\alpha_i(t)$
from Eqn.\ref{alpha_eqn} for the current time $t$ for all the states
$i$ and determine the most probable state at that time. So we can
describe the control output from the HMM at time $t$ given by
$h(t)\in\{1,..,m\}$ as
 \begin{equation}
   h(t)=\argmax_i\alpha_i(t).
 \label{forward_eqn}
\end{equation}
The output $h$ is used to determine the the control parameters
$(a(t),b(t))$ for aggregation and dispersion of the swarm, as
described in Sec.~\ref{Sec:FormationControl} below.

\subsection{Tracking Eye Movements}
In traditional EEG research, Eye Movement signals are considered as
artifacts and are removed.  In contrast, we use these signals as
inputs for our system to command the direction of travel for the
robots. There are various available methods in the literature for
detecting and tracking eye movements, which vary considerably
\cite{Eye_tracking_2007}. These methods can be broadly categorized
into (i) Contact based tracking which offer high accuracy and
sophistication, (ii) Non-contact based optical tracking methods which
measure relative positioning remotely with sensors such as cameras,
and (iii) measuring surface electrical potentials from skin, also
known as Electrooculogram (EOG), near the eyes.  Our EEG headset
detects these EOG signals related to eye movement, hence we can detect
eye movement with no additional hardware.

\subsubsection{EOG : An Overview}       
The human eye can be modeled as an electrical dipole whose axis is
roughly collinear to the axis of the human eye. The electrical dipole
rotates with the rotation of the eye causing small differences (in
microvolts) between the electrical potential at the skin surface
depending on eye position. The order of magnitude of these signals are
much larger than signals due to brain activity evident from
Fig. \ref{U_D_detection} and Fig.\ref{L_R_detection}, hence they can
be measured and contrasted easily.

\begin{figure}
\centering
\includegraphics[width=2in,height=2in,keepaspectratio]{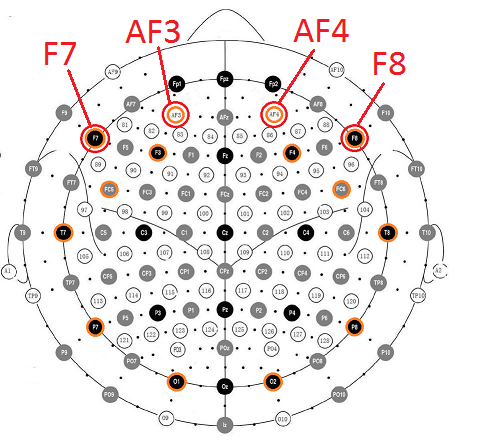}
\caption{EEG sensor placement on the human scalp using the 10-20
  system. The locations highlighted in orange depict the locations
  used by the Emotiv Epoc headset (\texttt{http://www.emotiv.com}). We
  specifically read the red circled locations to get EOG signals for
  eye movement. }
\label{sensor_placement}
\end{figure}

EOG typically uses exclusive electrodes around the eyes to measure
movements. But our electrode positions are fixed so we employ the four
closest electrodes to the eyes: `AF3',`AF4',`F7' and `F8', according
to the 10-20 EEG sensor placement system, as shown in the diagram in
Fig. \ref{sensor_placement}. Previous methods to detect eye motion
have relied on complex classification based algorithms.  In contrast,
our method uses a simple statistical calculation.

The spatio-temporal signals from these electrodes near the eyes can be
described by $u_i(t)\in\mathbb{R}$ where $i \in \{AF3,AF4,F7,F8\}$
denoting the electrodes used. We first normalize the signal by
subtracting its mean for each electrode to center the signals about
zero. This can be described
by $$u_i(t)-\frac{\sum_{t=0}^{\tau}u_i}{\tau},$$ where $\tau$ denotes
the number of samples used for the baseline removal.

\subsubsection{Horizontal Eye Movement Detection} 
Electrodes 'F7' and 'F8' are chosen for horizontal eye movement
detection as they are the farthest apart in the horizontal plane while
being closest to the eyes. Our algorithm for decoding horizontal
directional movement depicted in Fig. \ref{L_R_detection} is described
in Algorithm \ref{L_R_algo}.  In Fig. \ref{L_R_detection} the green
ellipses indicate the signal for leftward eye movement and the blue
ellipses indicate rightward eye movement. The red ellipse represents
blinks which are filtered out.
 
\begin{algorithm}
\caption{Horizontal Eye Movement Detection}\label{L_R_algo}
\begin{algorithmic}[1]
 \\Remove Baseline with $\tau=640$ samples.
 \\ Window the data with window size $w\in\mathbb{I^+}$
 \begin{equation}
 u_i^{\eta}(t)= \eta u_i(w) \label{Window}  
\end{equation}  
where $\eta$ represents the window number. We use $w=128$ samples
corresponding to 1 second of data with no overlap.  \\ Apply $8^{th}$
order 4 Hz low pass Butterworth filter to the windowed data to isolate
the eye movement signals.  \\ Subtract the resulting signals from both
electrodes \begin{equation}
  u_{F7-F8}^{\eta}(t)=u_{F7}^{\eta}(t)-u_{F8}^{\eta}(t)
 \end{equation} 
 \\ Detect peaks and troughs with threshold magnitude of $200 \mu V$
 and minimum seperation of $w-1$ samples in $u_{F7-F8}^{\eta}(t)$.  \\
 Assign Peaks to eye movements to the left $e_l^\eta\in\mathbb{Z^+}$
 and troughs to eye movements to the right $e_r^\eta\in\mathbb{Z^+}$.
\end{algorithmic}
\end{algorithm} 

\subsubsection{Vertical Eye movement Detection}
\begin{figure}
\centering
\includegraphics[width=3in,height=5in,keepaspectratio]{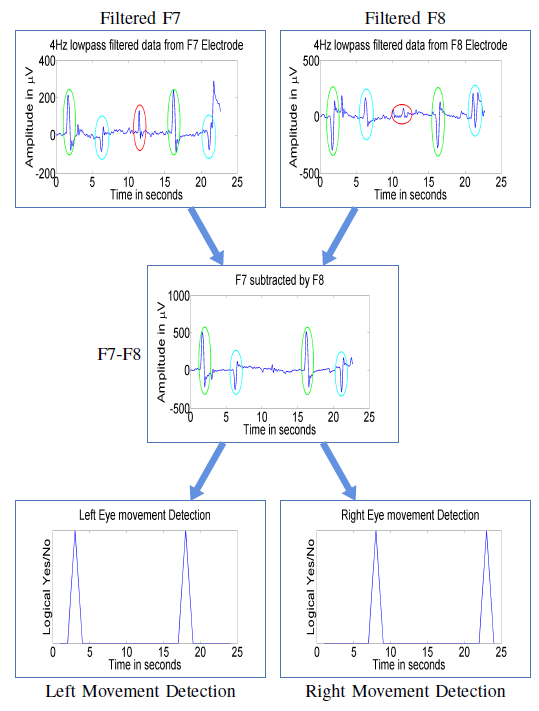}
\caption{Horizontal Eye Movement Tracking.  The main steps in
  Algorithm \ref{L_R_algo} are shown graphically from top to bottom.
  The bottom plots show the final extracted left and right eye
  movements.}
\label{L_R_detection}
\end{figure}

\begin{figure}
\centering
\includegraphics[width=3in,height=5in,keepaspectratio]{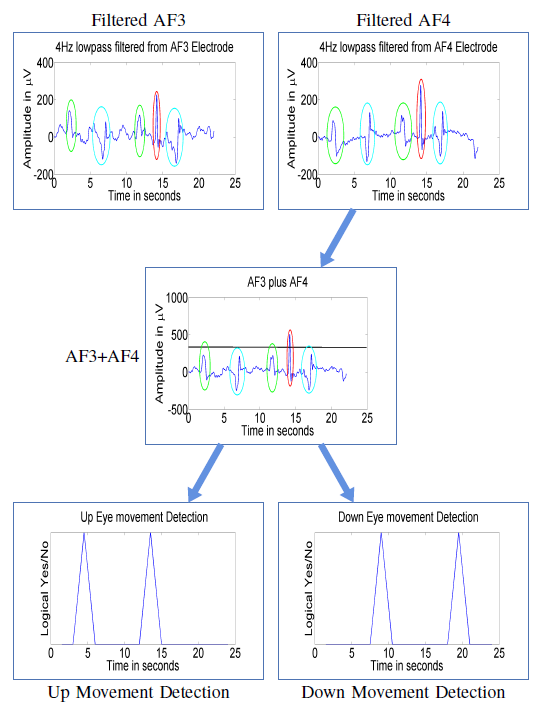}
\caption{Vertical Eye Movement Tracking. The main steps in Algorithm
  \ref{U_D_algo} are shown graphically from top to bottom.  The bottom
  plots show the final extracted up and down eye movements.}
\label{U_D_detection}
\end{figure}

Electrodes 'AF3' and 'AF4' are chosen for vertical eye movement
detection. The method used is different from horizontal eye movement
detection since we do not have any electrode below the eyes to detect
the dipoles in the vertical plane of the head. Eye movement upwards
results in a positive deflection in both electrodes whereas eye
movement downwards has negative deflection for both electrodes.  Our
algorithm for decoding vertical eye movement is depicted in Fig
\ref{U_D_detection} and described in Algorithm \ref{U_D_algo}. In Fig
\ref{U_D_detection}, the green ellipses indicate the signal for
vertical eye movement, and the blue ellipses indicate horizontal
movement. The red ellipse represents blinks which are filtered out.
\begin{algorithm}
\caption{Vertical Eye Movement Detection}\label{U_D_algo}
\begin{algorithmic}[1]
 \\ Remove Baseline with $\tau=640$ samples.
 \\ Window the data similar to Equation \ref{Window}
 \\ Same as step 3 in Algorithm 1.
 \\ Add the signals from both the electrodes to get \begin{equation}
 u_{AF3+AF4}^{\eta}(t)=u_{AF3}^{\eta}(t)+u_{AF4}^{\eta}(t)
 \label{AF3+AF4}
 \end{equation} 
 \\ Find peaks and troughs with minimum seperation of $w-1$ and a
 signal level of $150 \mu V$ - $250 \mu V$ from
 $u_{AF3+AF4}^{\eta}(t)$.  \\ Assign Peaks to upward eye movements
 $e_u^\eta\in\mathbb{Z^+}$ and troughs to downward eye movements
 $e_d^\eta\in\mathbb{Z^+}$.
\end{algorithmic}
\end{algorithm}
It should be noted that both the algorithms above also filter out
blinks, which look quite similar to the eye movements (see Figs.
\ref{U_D_detection} and \ref{L_R_detection}). In the case of
horizontal eye movements, the F7 and F8 electrodes both record blinks
with almost equal magnitude since they are located approximately at
the same distance from the eyes. Hence the signal
$u_{F7-F8}^{\eta}(t)$ is automatically devoid of blinks (Eqn
\ref{AF3+AF4}), as evident from Fig \ref{L_R_detection}.  For the
vertical eye movement detection, we introduce an upper threshold to
filter out the blinks, which typically are much larger than eye
movement signals, as can be seen in Fig. \ref{U_D_detection}. Hence,
our algorithm effectively discards blinks and measures only the user's
intentional eye movements.

Finally, after the left-right and up-down signals have been extracted
form the user's eye movements, these signals are used to control the
left-right and forward-backward motion of the robot swarm through the
control parameter $v(t)$, as described below in
Sec.~\ref{Sec:FormationControl}
 
%%%%%%%%%%%%%%%%%%%%%%%%%%% FIgures in Signal Processing %%%%%%%%%%%%%%%%%%%%%%%%%%

%%%%%%%%%%%%%%%%%%%%%%%%%%%%%%%%%%%%%%%%%%%%%%%%%%%%%%%%%%%%%%%%%%%%%%%%%%%%%%%%%%%

\subsection{Formation Control}
\label{Sec:FormationControl}
We described the general form of the potential field based formation
controller above in Sec~\ref{Sec:ProblemFormulation}.  We refer the
reader to
\cite{Olfati-Saber,Gazi&Passino,HowardEtAlDARS02PotentialFieldSensorNetDeployment}
for details and proofs about stability and convergence
properties. Here we describe how we control the size and motion of the
swarm through the parameters $(a, b, v)$.

\subsubsection{Controlling Size}
Recall that the system is described by the two dimensional state space
equation for the $i^{th}$ agent
\begin{equation}
  \dot{x_i} = \sum_{j=1}^M f_{ab}(x_i,x_j)+v.
\end{equation}
We let the interaction between robots $i$ and $j$ be given by
	  \begin{equation}
	  f_{ab}(x_i,x_j) =  \frac{a(x_j-x_i)}{(\|x_j-x_i\|-2r)^2} - \frac{b(x_j-x_i)}{(\|x_j-x_i\|-2r)^3}
	  \label{potential_function_eqn}
\end{equation}
Where $r$ is the radius of the robot.  We can see that
% $$g_{a}(\|y\|) = - \frac{a}{(\|x_j-x_i\|-2r)^2} , g_{r}(\|y\|) =
% \frac{b}{(\|x_j-x_i\|-2r)^3}$$ $g_a(\|x_j-x_i\|)$ and
% $g_r(\|x_j-x_i\|)$ are the attraction
% and repulsion components respectively which are based on
% the
% inter agent distances.
the left term provides the attracting field, and the right the
repelling field.  The $r$ term introduces a safety region around the
robots so collision can be avoided.

There is an equilibrium inter-robot distance for this system, in which
attraction and repulsion forces balance.  Let that equilibrium
distance be denoted $\delta$, so that
$g_a(\|\delta\|)=g_r(\|\delta\|)$. This $\delta$ is governed by the
attraction and repulsion gains $a$ and $b$ respectively.  In our
method we vary the gains to achieve different equilibrium
formation. We map the two state output from the HMM to two distinct
sets of gains in order to achieve the aggregation and dispersion of
the swarm.
	  
\subsubsection{Controlling size and motion}
Now to control the motion of the swarm we rely on the output from the
eye movement detection which gives us four possible motion commands:
Forward, Backward, Left, and Right. We use these command to assign
values to the vector $v$, which drives every robot in the swarm in the
same direction \ref{potential_function_eqn}.
% \begin{equation}
%   \dot{x_i} = \sum_{j=1}^M f_{ij}(x_i,x_j)+v,
% \end{equation}
Depending on the eye movement the vector $v$ is assigned preset values
which makes all the agents in the swarm move locally in the direction
of $v$ independent of the swarms aggregation or dispersion. 
 
\section{SIMULATIONS AND HARDWARE EXPERIMENTS}
\label{Sec:Experiments}
\begin{figure}
\centering
\includegraphics[width=2.25in,height=2.25in,keepaspectratio]{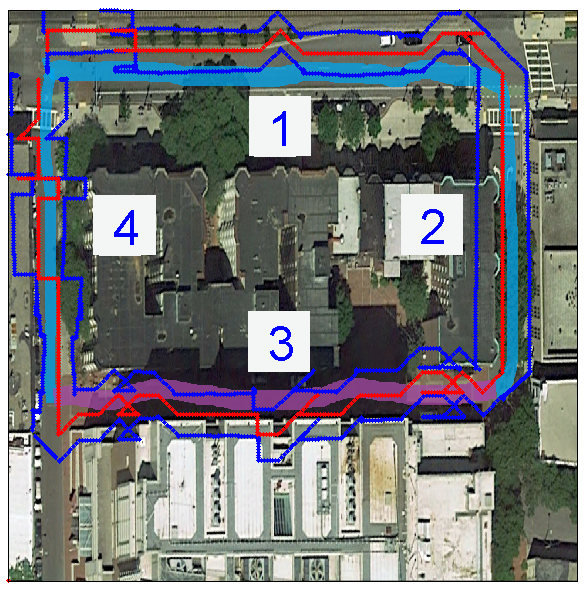}
\caption{Path of the swarm (blue) along with its centroid (red) traveling in a simulated environment with 128 robots.}
\label{centroid_sim}
\end{figure}

To demonstrate our brain-swarm interface, we developed a simulation
environment in Matlab.  We chose a section of the Boston University
campus, with a rectangular path around a campus building, as shown in
\ref{centroid_sim}. The path is divided into 4 edges and the swarm has
to be driven starting from the left of edge 1 and end on the top of
edge 4 following a clockwise motion. At edge 3 (purple path) due to
the narrow passageway, the user has to make the swarm aggregate into a
tighter swarm by switching thoughts, while in edges 1, 2 and 4 (Blue
path) the user makes the swarm disperse.

For the training phase, the user switched between two thoughts at
least twice over a period of 60 seconds, during which the EEG signals
were recorded and fed into the Baum-Welch Algorithm
(Fig. \ref{Fig:Baum_welch}) to get the model parameters as described
in Sec.~\ref{Sec:HMM}. The thoughts used for simulations and
experiment were distinct and repeatable: the disperse state was
invoked with a relaxed neutral thought, while the aggregate state was
invoked by a mentally challenging task (in this case calculating the
Fibonacci series). Then EEG signals were streamed live and processed
as discussed previously to generate the control inputs $\Theta$.  

The simulated swarm consisted of 128 point sized holonomic robots. The
attraction gain $a$ was fixed at $1$ and the repulsion gain $b$ was
calculated according to $b(t)=h(t)*M/2.625$, where $h(t)$ is the
estimated state sequence from (\ref{forward_eqn}). In our $2$ state
HMM case according to the previous formula the user's thought
corresponding to state $1$ causes the robots to disperse and increase
swarm size, and for state $2$ causes the robots to converge and
aggregate to a smaller size. The swarm reaches its equilibrium size
for a particular thought state and stays at that size until the user
switches thoughts. 

\begin{figure}
\centering
\subfigure[State estimation using $\alpha$ during simulation\label{alpha_sim}]{\includegraphics[width=0.49\linewidth,keepaspectratio]{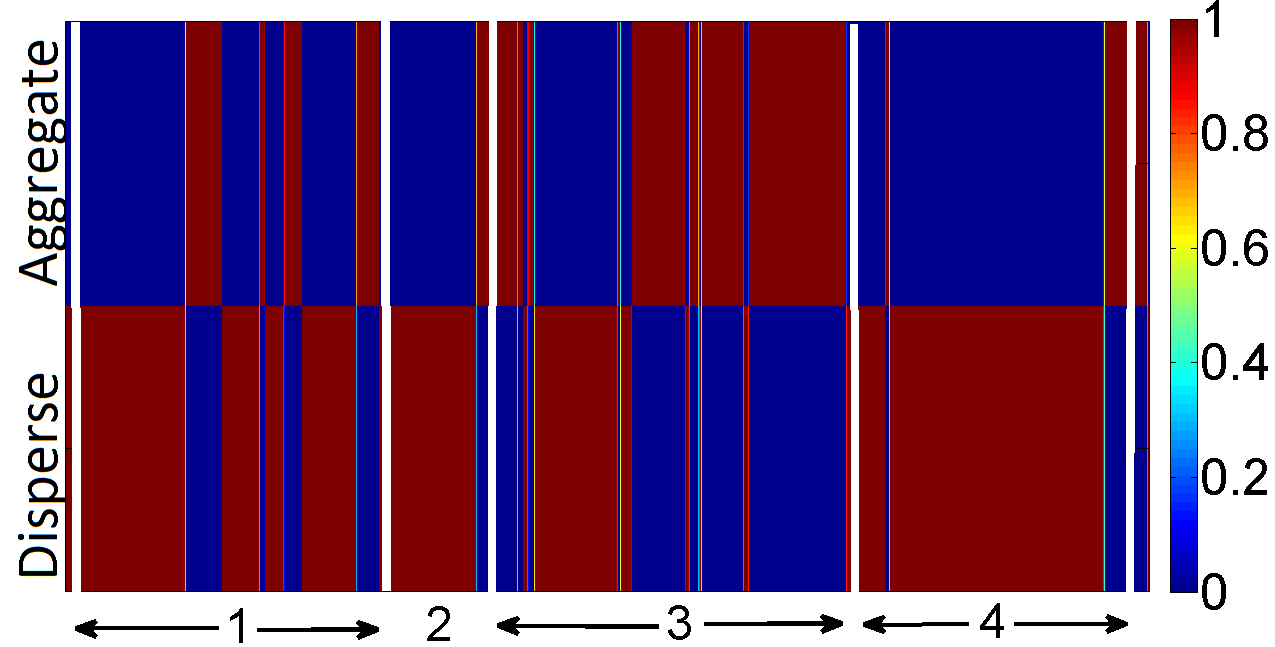}}
\subfigure[Horizontal Eye movement estimation\label{LR_sim}]{\includegraphics[width=0.49\linewidth,keepaspectratio]{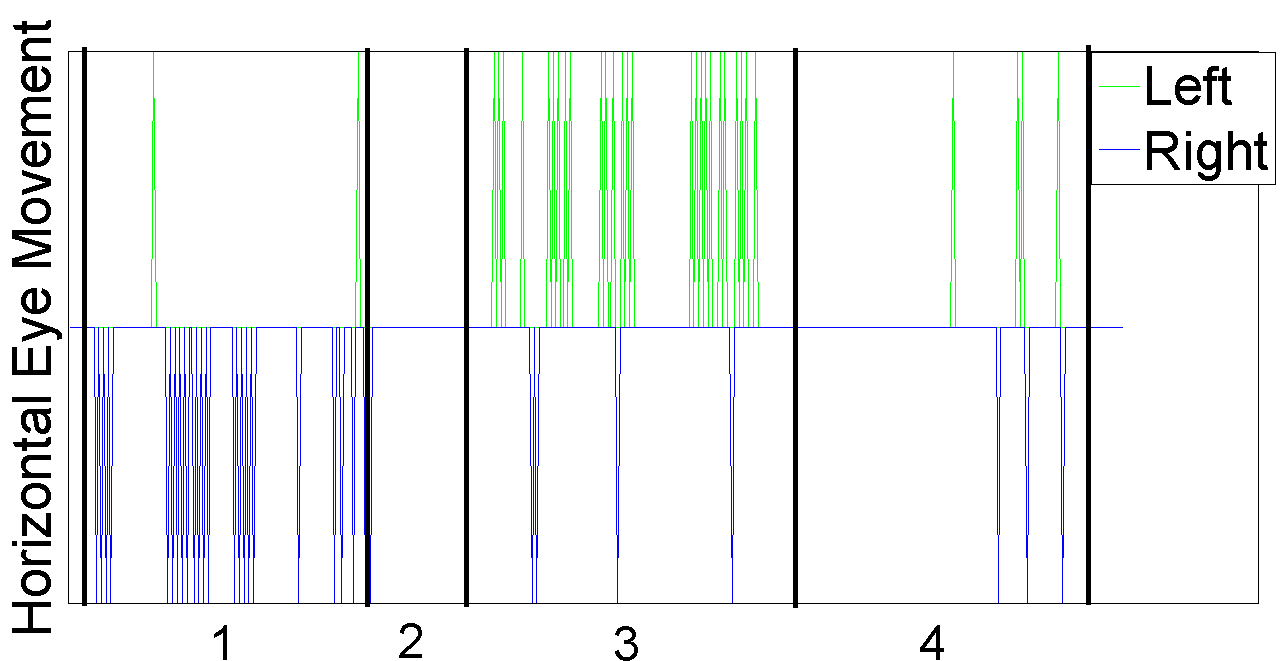}}
\subfigure[Vertical Eye movement estimation\label{UD_sim}]{\includegraphics[width=0.49\linewidth,keepaspectratio]{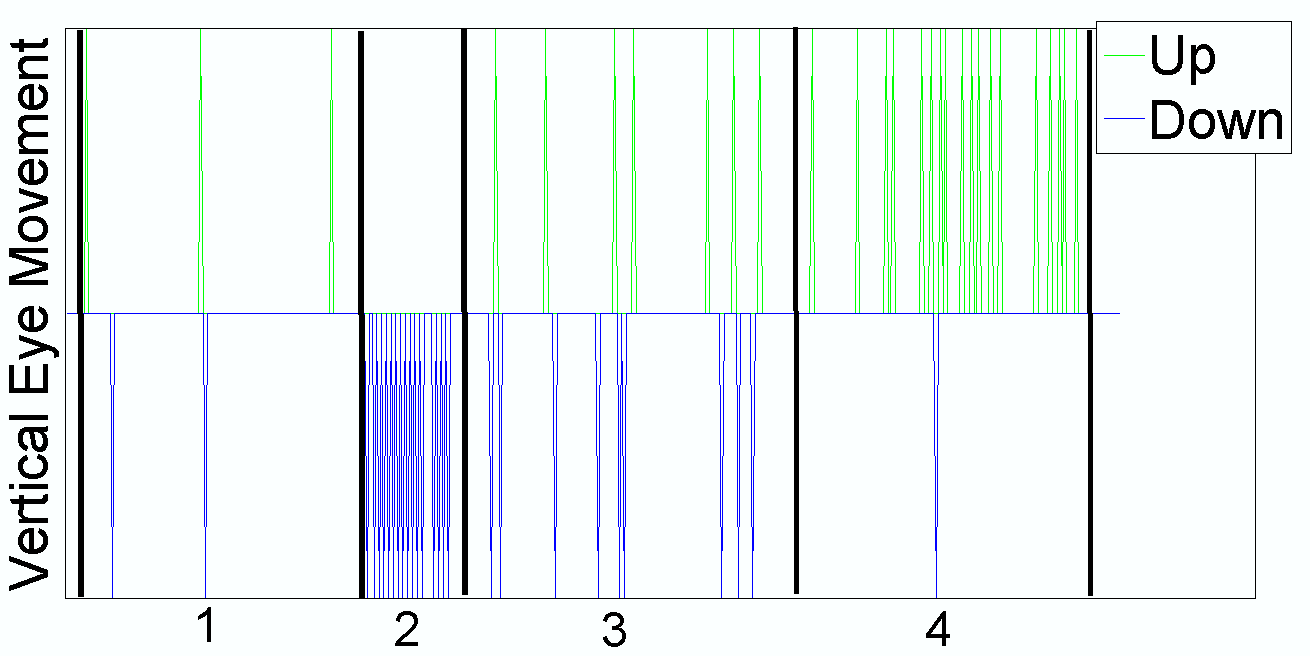}}
\caption{Simulation Results. a) Shows the color map of the variable
  $\alpha$ during the forward procedure. b) Shows the horizontal eye
  movement. c) Shows the vertical eye movement. \label{Sim_results}}
\end{figure}

The results of the simulation exercise is summarized in
Figs.~\ref{Sim_results} and \ref{centroid_sim}.
Fig.~\ref{Sim_results} shows the time history of eye movement
detection and the mental thought estimation during the motion of the
swarm along the 4 legs of the path. It can be seen that the thought
estimation remains mostly in the disperse state during legs 1, 2, and
4, and is mostly in the aggregate state in Sector 3, as intended.  Minor
inaccuracies can be attributed not only to the stochastic nature of
the HMM, but also to the quality of user's thoughts and noise in the
EEG headset.  From Fig.~\ref{centroid_sim} which shows the path of the
centroid of the swarm, we can see the Eye Movement detection is
successful in steering the swarm, with a few misclassifications due to
the nature of the noisy EEG signals and non-intentional eye movements.
We refer the reader to the accompanying video.

\begin{figure}
\centering
\includegraphics[width=2.5in,height=1in,keepaspectratio]{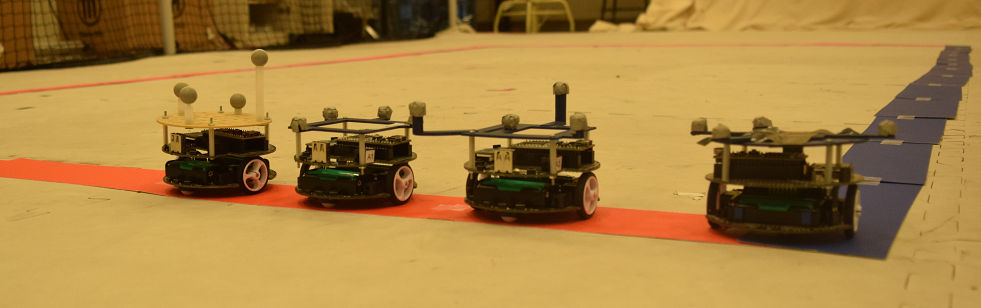}
\caption{M3pi Robots in the Optitrack Arena.}
\label{M3pi_Robots}
\end{figure}

\begin{figure}
\centering
\includegraphics[width=2.1in,height=2.1in,keepaspectratio]{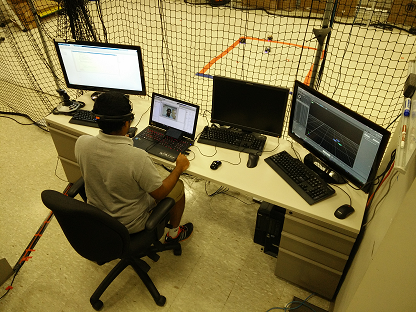}
\caption{Experimental area and setup.}
\label{exp_setup}
\end{figure}

\begin{figure}
\centering
\includegraphics[width=2.25in,height=2.25in,keepaspectratio]{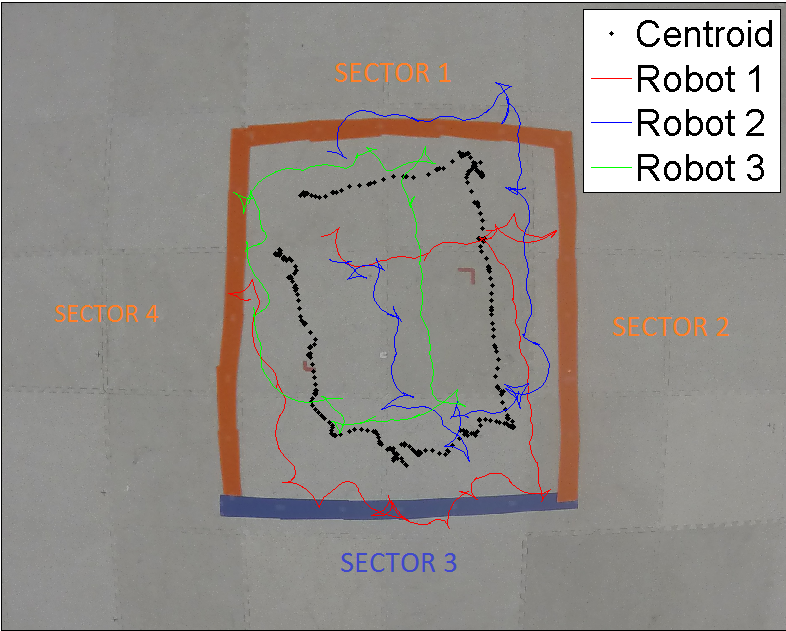}
\caption{Position of the swarm of 3 robots along with their centroid
  during the experiment.}
\label{centroid_exp}
\end{figure}

For the hardware experiments we used the m3pi platform with an Mbed
controller for mobile swarming robots, and Zigbee radios for
communication.  The experiments were carried out in an environment
with an Optitrack motion capture system to track the motion of the
robots (see Fig.~\ref{M3pi_Robots}).  The control parameters used were
$a = 4$ and $b = 80$ for aggregation, and $a = 2$ and $b = 80$ for
dispersion. These two wheeled differential drive robots receive
individual motor speeds as control inputs from the computer. A
proportional point-offset controller is used to generate the motor
speeds from the potential function controller (the details of which
can be found in \cite{Alyssa_poc}). The control commands for the
robots were computed off board the robots, and set to the robots over
Zigbee at an update rate of 30 Hz. Due to computational and hardware
complexities the computations were divided among 3 computers (one for
Optitrack data acquisition, one for controller implementation, and
another other for EEG signal processing and video recording) as shown
in the experimental setup in Fig.~\ref{exp_setup}.  The tcp/ip
protocol was used for communication among them.  The experimental area
(Fig.~\ref{centroid_exp} ) was chosen to be a rectangular area divided
into 4 legs, similarly to the simulation.  In sector 3 the user again
must make the swarm aggregate, and in the other legs the swarm should
disperse, while navigating in a clockwise manner starting from sector
1.  The user obtained visual feedback of the position of the swarm by
viewing a live feed of the experimental area from a GoPro camera on a
mobile device.

\begin{figure}
\centering
\subfigure[State estimation using $\alpha$ during experiment\label{alpha_exp}]{\includegraphics[width=0.49\linewidth,keepaspectratio]{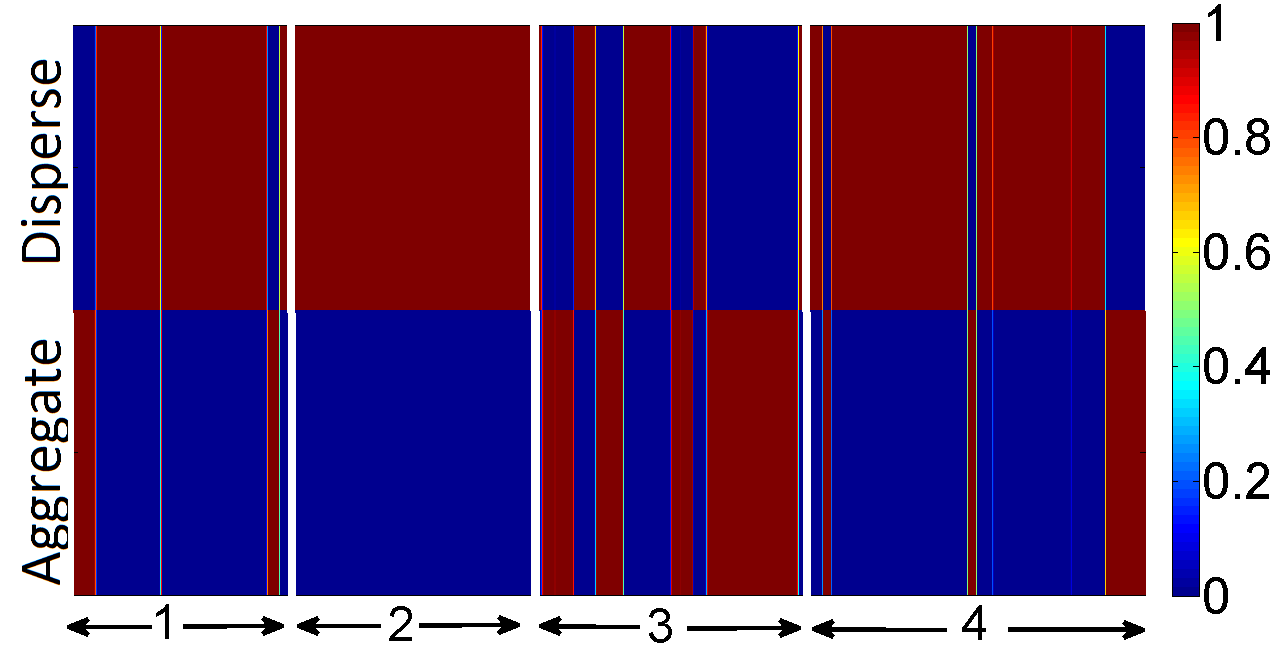}}
\subfigure[Horizontal Eye movement estimation\label{LR_exp}]{\includegraphics[width=0.49\linewidth,keepaspectratio]{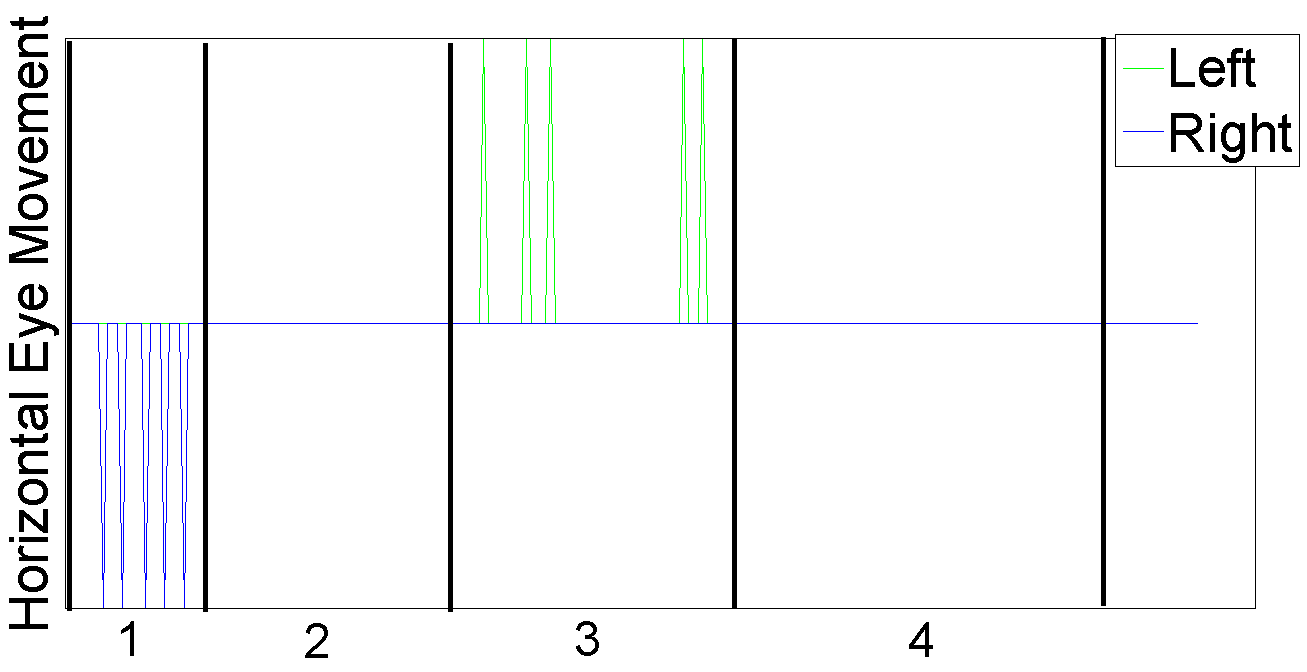}}
\subfigure[Vertical Eye movement estimation\label{UD_exp}]{\includegraphics[width=0.49\linewidth]{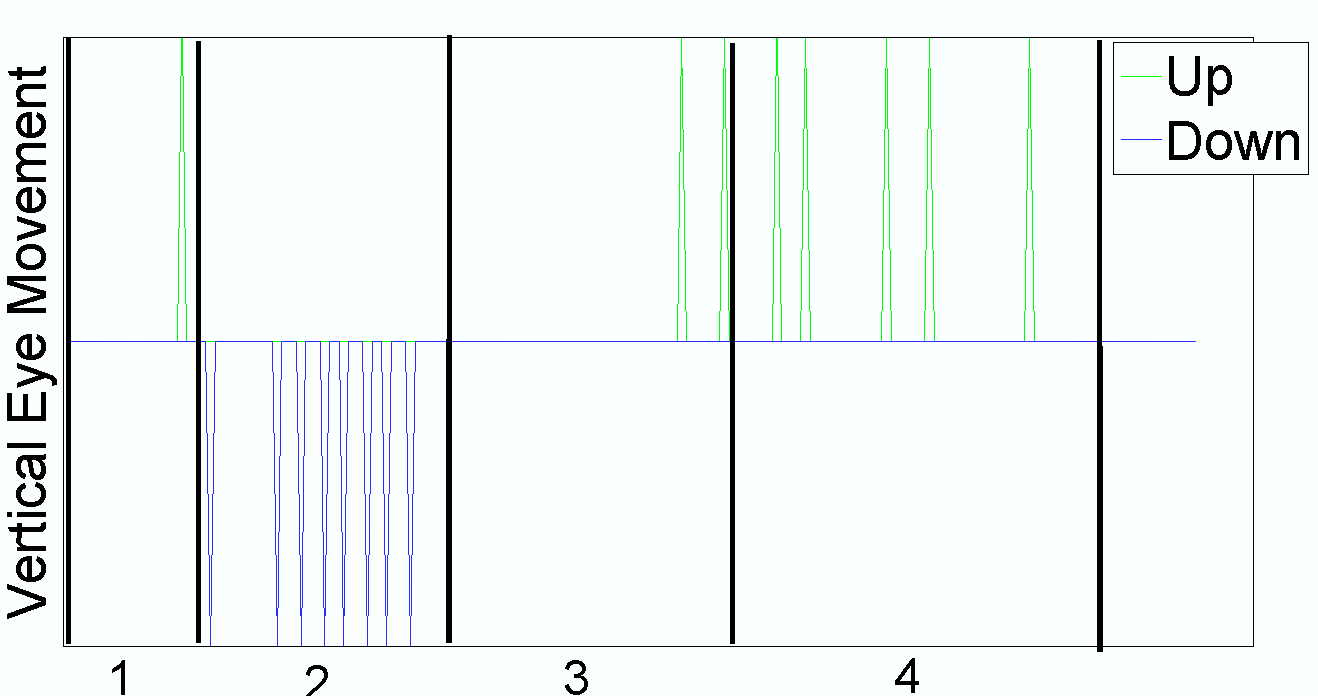}}
\subfigure[Distances from the swarm centroid\label{cent_dist_exp}]{\includegraphics[width=0.49\linewidth]{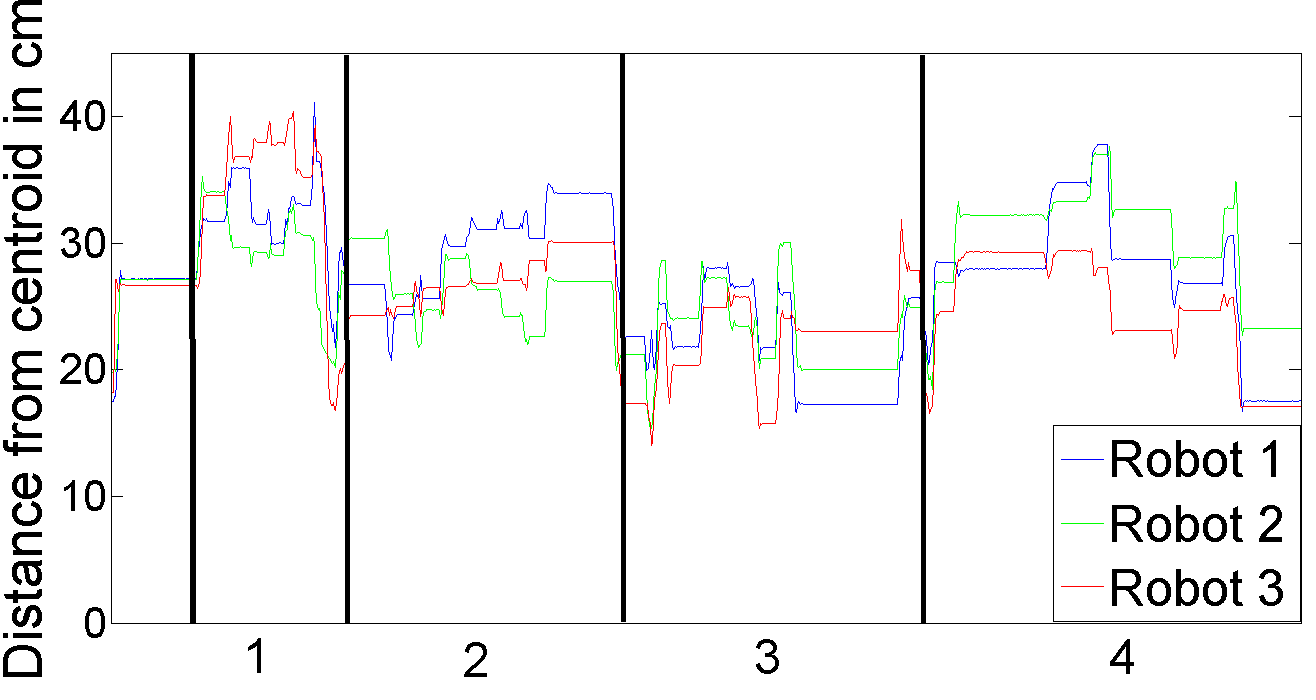}}
\caption{Experiment Results. a) Shows the color map of the variable
  $\alpha$ during the forward procedure. b) Shows the horizontal eye
  movement. c) Shows the vertical eye movement. d) Shows the distance
  from centroid of the three robots in centimeter.}
\label{Exp_result}
\end{figure}

The experimental results are summarized in Figs.~\ref{Exp_result} and
\ref{centroid_exp}. Fig.~\ref{Exp_result} shows the time history of
eye movement detection and the thought state estimation throughout the
motion of the swarm along the 4 legs. From Fig.~\ref{centroid_sim}
which shows the path of the robots and their centroid, we can see the
eye movement detection is successful in steering the swarm. The color
map for the thought state estimation again shows that the HMM is able
to reliably determine the user's intention.  The system remains in the
disperse state with high confidence during legs 1, 2, and 4, and in
the aggregate state with high confidence during sector 3.  We again refer
the reader to the accompanying experiment video.

The biggest challenge in this work is in the integration of this
complex system with interacting hardware, communication, software, and
human components. We used holonomic dynamics during simulation whereas
The M3pis are nonholonomic robots with inefficient actuation and
communication.  In addition, it was quite a mental challenge for the
user to concentrate on thoughts, eye movement, and system monitoring
simultaneously. Despite these challenges, we were able to successfully
demonstrate the proposed method.

\section{CONCLUSIONS}
\label{Sec:Conclusions}
In this paper we propose and successfully demonstrate an online Brain
Swarm Interface to control a swarm of ground vehicles in simulation
and experiments using off-the-shelf hardware. We integrate a variety
of engineering and scientific techniques in neuroscience, signal
processing, machine learning, control theory, and swarm robotics to
construct and implement our system. We successfully navigate a robot
swarm in simulation and experiment on a given path. The techniques
developed in this paper are a proof of concept to demonstrate that
a swarm of robots can be controlled by the thoughts and eye movements
of a human user.  These techniques can be applied to create an intuitive
interface especially for people to control multiple objects
in their environment simultaneously.  However, there is significant
room for improvement.  In the future we plan to develop a richer HMM
algorithm that can detect a larger range of user intentions for the
swarm, including the ability to induce a range of different shapes and
motions for the swarm.  We also plan to extend the algorithm to
control a swarm of quadrotors robots in 3 dimensional space.

\section*{ACKNOWLEDGMENTS}
We thank Alyssa Pierson for her help in developing the point offset
controller to control individual M3pi robots.

% \bibliographystyle{IEEEtran}
% \bibliography{reference}

\end{document}